\documentclass{article}
\usepackage{nips13submit_e,times}
\usepackage{hyperref}
\usepackage{url}
\usepackage{microtype}
\usepackage{booktabs}
\usepackage{amssymb,amsmath,graphicx,color,framed,soul}
\usepackage{siunitx}
\usepackage[numbers]{natbib}
\newcommand{\citeg}[1]{\citep[e.g.][]{#1}}
\usepackage[small]{bibhacks}
\usepackage{algorithm}
\usepackage{algpseudocode}

\newcommand{\eg}{e.g.\ }
\newcommand{\etal}{\textit{et al}.\ }
\newcommand{\realdomain}{\mathbb{R}}
\newcommand{\given}{\!\mid\!}

\renewcommand{\vec}[1]{\boldsymbol{#1}}
\newcommand{\mat}[1]{\boldsymbol{#1}}

\newcommand{\transp}{^{\top}}

\newcommand{\sigm}{\text{sigm}}
\newcommand{\softmax}{\text{softmax}}

\newcommand{\xltd}{\vec{x}_{<d}}
\newcommand{\hd}{\vec{h}_d}
\newcommand{\vd}{\vec{v}_d}

\title{RNADE: The real-valued neural autoregressive density-estimator}

\author{
Benigno Uria \ \mbox{\rm and} \ Iain Murray\\
School of Informatics\\
University of Edinburgh\\
\texttt{\{b.uria,i.murray\}@ed.ac.uk} \\
\And
Hugo Larochelle\\
D{\'e}partement d'informatique\\
Universit{\'e} de Sherbrooke\\
\texttt{hugo.larochelle@usherbrooke.ca} \\
}

\nipsfinalcopy

\begin{document}
\maketitle

\begin{abstract}

We introduce RNADE, a new model for joint density estimation of real-valued
vectors. Our model calculates the density of a datapoint as the product of
one-dimensional conditionals modeled using mixture density networks with shared
parameters. RNADE learns a distributed representation of the data, while having
a tractable expression for the calculation of densities. A tractable likelihood
allows direct comparison with other methods and training by standard
gradient-based optimizers. We compare the performance of RNADE on several
datasets of heterogeneous and perceptual data, finding it outperforms mixture
models in all but one case.

\end{abstract}

\section{Introduction}

Probabilistic approaches to machine learning involve modeling the
probability distributions over large collections of variables. The number
of parameters required to describe a general discrete distribution grows
exponentially in its dimensionality, so some structure or regularity must
be imposed, often through graphical models \citep[e.g.][]{koller2009}.
Graphical models are also used to describe probability densities over
collections of real-valued variables.

Often parts of a task-specific probabilistic model are hard to specify, and
are learned from data using generic models. For example, the natural
probabilistic approach to image restoration tasks (such as denoising,
deblurring, inpainting) requires a multivariate distribution over
uncorrupted patches of pixels. It has long been appreciated that large
classes of densities can be estimated consistently by kernel density
estimation \citep{cacoullos1966}, and a large mixture of Gaussians can
closely represent any density. In practice, a parametric mixture of
Gaussians seems to fit the distribution over patches of pixels and obtains
state-of-the-art restorations \citep{Zoran2011}.
It may not be possible to fit small image patches significantly better, but
alternative models could further test this claim. Moreover, competitive
alternatives to mixture models might improve performance in other applications
that have insufficient training data to fit mixture models well.

Restricted Boltzmann Machines (RBMs), which are undirected graphical
models, fit samples of binary vectors from a range of sources better than
mixture models \citep{Salakhutdinov2008a,Larochelle2011}. One explanation
is that RBMs form a distributed representation: many hidden units are
active when explaining an observation, which is a better match to most real
data than a single mixture component. Another explanation is that RBMs
\emph{are} mixture models, but the number of components is exponential
in the number of hidden units. Parameter tying among components
allows these more flexible models to generalize better from small numbers
of examples. There are two practical difficulties with RBMs: the likelihood
of the model must be approximated, and samples can only be drawn from the
model approximately by Gibbs sampling. The Neural Autoregressive Distribution
Estimator (NADE) overcomes these difficulties~\cite{Larochelle2011}. NADE is a
directed graphical model, or feed-forward neural network, initially derived as
an approximation to an RBM, but then fitted as a model in its own right.

In this work we introduce
the Real-valued Autoregressive Density Estimator (RNADE),
an extension of NADE\@. An autoregressive model
expresses the density of a vector as an ordered product of one-dimensional
distributions, each conditioned on the values of previous dimensions in the
(perhaps arbitrary) ordering. We use the parameter sharing previously
introduced by NADE, combined with mixture density networks
\citep{Bishop1994}, an existing flexible approach to modeling real-valued
distributions with neural networks. By construction, the density of a test
point under RNADE is cheap to compute, unlike RBM-based models. The
neural network structure provides a flexible way to alter the mean and
variance of a mixture component depending on context, potentially modeling
non-linear or heteroscedastic data with fewer components than unconstrained
mixture models.

\section{Background: Autoregressive models}

Both NADE~\citep{Larochelle2011} and our RNADE model are based on the chain
rule (or product rule), which factorizes any distribution over a vector of
variables into a product of terms: $p(\vec{x})=\prod_{d=1}^{D}p(x_d \given
\xltd)$, where $\xltd$ denotes all attributes preceding $x_d$ in a fixed
arbitrary ordering of the attributes. This factorization corresponds to
a Bayesian network where every variable is a parent of all variables after it%
.
As this model assumes no conditional independences, it says nothing about
the distribution in itself. However, the (perhaps arbitrary)
ordering we choose will matter if the form of the conditionals
is constrained. If we assume tractable parametric forms for
each of the conditional distributions, then the joint distribution can be
computed for any vector, and the parameters of the model can be locally
fitted to a penalized maximum likelihood objective using any gradient-based
optimizer.

For binary data, each conditional distribution can be modeled with logistic
regression, which is called a fully visible sigmoid belief network
(FVSBN)~\citep{Frey1996}. Neural networks can also be used for each binary
prediction task~\citep{Bengio2000}. The neural autoregressive
distribution estimator (NADE) also uses neural networks for each
conditional, but with parameter sharing inspired by a mean-field approximation
to Restricted Boltzmann Machines~\citep{Larochelle2011}.
In detail, each conditional is given by a feed-forward
neural network with one hidden layer, $\hd\in\realdomain^H$:
\begin{equation}
p(x_d=1 \given \xltd)=
\sigm\left(\vd\transp\hd + b_d\right) \quad \mathrm{where} \quad
\hd=\sigm\left(\mat{W}_{\cdot,<d}\xltd + \vec{c}\right)\mathrm{,}
\label{eqn:nade}
\end{equation}
where $\vd \!\in\! \realdomain^H$, $b_d \!\in\! \realdomain$, $\vec{c} \!\in\! \realdomain^H$,
and $\mat{W} \!\!\in\! \realdomain^{H\times(D-1)}$ are neural network
parameters, and $\sigm$ represents the logistic sigmoid function
$1/(1+e^{-x})$.

The weights between the inputs and the hidden units for each neural network are
tied: $\mat{W}_{\cdot,<d}$ is the first $d\!-\!1$ columns of a shared weight
matrix $\mat{W}$. This parameter sharing reduces the total number of parameters
from quadratic in the number of input dimensions to linear, lessening the need
for regularisation. Computing the probability of a datapoint can also be done
in time linear in dimensionality, $O(DH)$, by sharing the computation when
calculating the hidden activation of each neural network
($\vec{a}_d=\mat{W}_{\cdot,<d}\xltd + \vec{c}$):
\begin{equation}
\vec{a}_1 = \vec{c}, \qquad
\vec{a}_{d+1} = \vec{a}_d+x_d\vec{W}_{\cdot,d}. \label{eq:tied-activations}
\end{equation}
When approximating Restricted Boltzmann Machines, the output weights
$\{\vd\}$ in \eqref{eqn:nade} were originally tied to the input
weights~$\mat{W}$. Untying these weights gave better statistical performance on a
range of tasks, with negligible extra computational cost~\citep{Larochelle2011}.

NADE has recently been extended to count data~\citep{larochelleneural}. The
possibility of extending generic neural autoregressive models to continuous data
has been mentioned~\citep{Bengio2000,bengiodiscussion}, but has not been
previously explored to our knowledge. An autoregressive mixture of experts with
scale mixture model experts has been developed as part of a sophisticated multi-resolution
model specifically for natural images~\citep{Theis2012a}. In more general work, Gaussian processes have been
used to model the conditional distributions of a fully visible Bayesian
network~\cite{Friedman2000}. However, these `Gaussian process networks' cannot
deal with multimodal conditional distributions or with large datasets (currently
$\gtrapprox 10^4$ points would require further approximation). In the next
section we propose a more flexible and scalable approach.

\section{Real-valued neural autoregressive density estimators}

The original derivation of NADE suggests deriving a real-valued version from a
mean-field approximation to the conditionals of a Gaussian-RBM\@. However, we
discarded this approach because the limitations of the Gaussian-RBM are well
documented~\cite{murray2009,Theis2011}: its isotropic conditional noise
model does not give competitive density estimates.
Approximating a more capable RBM model, such as the
mean-covariance~RBM~\cite{Ranzato2010} or the
spike-and-slab~RBM~\cite{Courville2011}, might be a fruitful future direction.

The main characteristic of NADE is the tying of its input-to-hidden weights. The
output layer was `untied' from the approximation to the RBM to give the model
greater flexibility. Taking this idea further, we add more parameters to NADE
to represent each one-dimensional conditional distribution with a mixture of
Gaussians instead of a Bernoulli distribution. That is, the outputs are mixture
density networks~\cite{Bishop1994}, with a shared hidden layer, using the same
parameter tying as NADE\@.

Thus, our \emph{Real-valued Neural Autoregressive Density-Estimator} or RNADE
model represents the probability density of a vector as:
\begin{equation}\label{eq:NN_FVBN}
p(\vec{x})=\prod_{d=1}^{D}p(x_d \given \xltd) \mathrm{\quad with \quad}
p(x_d \given \xltd)=p_\mathcal{M}(x_d \given \vec{\theta}_d)\mathrm{,}
\end{equation}
where $p_\mathcal{M}$ is a mixture of Gaussians with
parameters~$\vec{\theta}_d$. The mixture model parameters are calculated using a
neural network with all of the preceding dimensions, $\xltd$, as inputs. We now
give the details.

RNADE computes the same hidden unit activations, $\vec{a}_d$, as before using%
~\eqref{eq:tied-activations}.
As discussed by \citet{bengiodiscussion}, as an RNADE (or a NADE) with
sigmoidal units progresses across the input dimensions $d \in \{1 \ldots D\}$,
its hidden units will tend to become more and more saturated, due to their input
being a weighted sum of an increasing number of inputs. Bengio proposed
alleviating this effect by rescaling the hidden units' activation by a
free factor $\rho_d$ at each step, making the hidden unit values
\begin{equation}
    \hd=\sigm\left(\rho_d\vec{a}_d \right).
\label{eqn:rescaled-hidden-activations}
\end{equation}
Learning these extra rescaling parameters worked slightly better, and
all of our experiments use them.

Previous work on neural networks with real-valued outputs has found that
rectified linear units can work better than sigmoidal non-linearities
\citep{nair2010}. The hidden values for rectified linear units are:
\begin{equation}
\hd=\begin{cases}
    \rho_d\vec{a}_d & \text{if $\rho_d\vec{a}_d > 0$} \\
    0 & \text{otherwise.}
\end{cases}
\label{eqn:rescaled-rlu-hidden-activations}
\end{equation}
In preliminary experiments we found that these hidden units worked better than
sigmoidal units in RNADE, and used them throughout (except for an example
result with sigmoidal units in Table~\ref{tab:BSDS-results}).

Finally, the mixture of Gaussians parameters
for the $d$-th conditional, $\vec{\theta}_d=\left\{\vec{\alpha}_d, \vec{\mu}_d, \vec{\sigma}_d \right\} $, are set by:
\begin{align}
    \text{$K$ mixing fractions,} & & \hspace*{-1cm} \vec{\alpha}_d & =
    \softmax\left({\mat{V}_d^\alpha}\transp\hd + \vec{b}_d^\alpha\right) \\[-0.03in] \text{$K$ component means,} & & \hspace*{-1cm} \vec{\mu}_d & = {\mat{V}_d^\mu}\transp\hd + \vec{b}_d^\mu \\[-0.03in]
    \text{$K$ component standard deviations,} & & \hspace*{-1cm} \vec{\sigma}_d & = \exp\left({\mat{V}_d^\sigma}\transp\hd + \vec{b}_d^\sigma\right),
\end{align}
where free parameters $\mat{V}_d^\alpha$, $\mat{V}_d^\mu$, $\mat{V}_d^\sigma$ are $H \!\times\! K$
matrices, and $\vec{b}_d^\alpha$, $\vec{b}_d^\mu$, $\vec{b}_d^\sigma$ are
vectors of size $K$.
The \textit{softmax}~\cite{Bridle1989} ensures the mixing fractions are positive
and sum to one, the exponential ensures the standard deviations are positive.

Fitting an RNADE can be done using gradient ascent on the model's likelihood
given a training set of examples. We used minibatch stochastic gradient ascent in all
our experiments. In those RNADE models with MoG conditionals, we multiplied the
gradient of each component mean by its standard deviation (for a Gaussian,
Newton's method multiplies the gradient by its variance, but empirically
multiplying by the standard deviation worked better). This gradient scaling makes tight components
move more slowly than broad ones, a heuristic that we found allows the use of
higher learning rates.

\textbf{Variants:~~} Using a mixture of Gaussians to represent the conditional
distributions in RNADE is an arbitrary parametric choice. Given several
components, the mixture model can represent a rich set of skewed and multimodal
distributions with different tail behaviors. However, other choices could be
appropriate in particular circumstances. For example, work on natural images
often uses scale mixtures, where components share a common mean. Conditional
distributions of perceptual data are often assumed to be Laplacian \citeg{robinson1994}.
We call our main variant with mixtures of Gaussians RNADE-MoG, but also
experiment with mixtures of Laplacian outputs, RNADE-MoL\@.

\section{Experiments}

We compared RNADE to mixtures of Gaussians~(MoG) and factor analyzers~(MFA),
which are surprisingly strong baselines in some
tasks~\citep{Tang2012,Zoran2012}. Given the known poor performance of discrete
mixtures~\citep{Salakhutdinov2008a,Larochelle2011}, we limited our experiments
to modeling continuous attributes. However it would be easy to include both
discrete and continuous variables in a NADE-like architecture.

\subsection{Low-dimensional data}
We first considered five UCI datasets~\citep{Bache+Lichman:2013}, previously
used to study the performance of other density
estimators~\cite{Silva2011,Tang2012}. These datasets have relatively low
dimensionality, with between 10 and~32 attributes, but have hard thresholds and
non-linear dependencies that may make it difficult to fit mixtures of Gaussians
or factor analyzers.

\begin{table}
\vspace*{-0.1in}
\begin{center}
\caption{Average test-set log-likelihood per datapoint for 4 different models
on five UCI datasets. Performances not in bold can be shown to be significantly
worse than at least one of the results in bold as per a paired $t$-test on the
ten mean-likelihoods, with significance level 0.05.}
\label{tab:UCI-results}
\medskip
{%
\begin{tabular}{lrrSSccc}
\toprule
{Dataset} & {dim} & {size} & {Gaussian} & {\small MFA} & {\small
FVBN} & {\small RNADE-MoG} & {\small RNADE-MoL}\\
\midrule
Red wine & 11 & 1599 & -13.18 & -10.19 & \hspace{-0.5ex}$-11.03$ &
$\mathbf{-9.36}$ & \kern6pt$\mathbf{-9.46}$\\
White wine & 11 & 4898 & -13.20 & -10.73 & \hspace{-0.5ex}$-10.52$ &
\kern1pt$\mathbf{-10.23}$\;\; & $-10.38$ \\
Parkinsons & 15 & 5875 & -10.85 & -1.99 &
\hspace{-0.5ex}\kern6pt$\mathbf{-0.71}$ & $\mathbf{-0.90}$ & \kern5pt$-2.63$\\
Ionosphere & 32 & 351 & -41.24 & -17.55 & \hspace{-0.5ex}$-26.55$ &
$\mathbf{-2.50}$ & \kern6pt$\mathbf{-5.87}$\\
Boston housing\hspace{-0.5ex} & 10 & 506 & -11.37 & -4.54 &
\hspace{-0.5ex}\kern5pt$\mathbf{-3.41}$ & $\mathbf{-0.64}$ & \kern5pt$-4.04$
\\
\bottomrule
\end{tabular}}
\end{center}
\vspace*{-0.05in}
\end{table}

Following
Tang~\etal\cite{Tang2012}, we eliminated
discrete-valued attributes and an attribute from every
pair with a Pearson correlation coefficient greater than~0.98. Each dimension of
the data was normalized by subtracting its training subset sample mean and
dividing by its standard deviation. All results are reported on the normalized
data.

As baselines we fitted full-covariance Gaussians and mixtures of factor analysers.
To measure the performance of the different models, we calculated their
log-likelihood on held-out test data. Because these datasets are small,
we used 10-folds, with 90\% of the data for training, and 10\% for testing.

We chose the hyperparameter values for each model by doing per-fold
cross-validation; using a ninth of the training data as validation data. Once
the hyperparameter values had been chosen, we trained each model using all the
training data (including the validation data) and measured its performance on
the 10\% of held-out testing data. In order to avoid overfitting, we stopped the
training after reaching a training likelihood higher than the one obtained on
the best validation-wise iteration of the corresponding validation run. Early
stopping is crucial to avoid overfitting the RNADE models. It also improves the
results of the MFAs, but to a lesser degree.

The MFA models were trained using the EM algorithm~\cite{Ghahramani1996,verbeek2005}, the
number of components and factors were crossvalidated. The number of factors was
chosen from even numbers from $2\dots D$, where selecting $D$ gives a mixture of Gaussians. The
number of components was chosen among all even numbers from $2\dots 50$
(crossvalidation always selected fewer than $50$ components).

RNADE-MoG and RNADE-MoL models
were fitted using minibatch stochastic gradient descent, using
minibatches of size 100, for 500 epochs, each epoch comprising 10 minibatches.
For each experiment, the number of hidden units (50), the non-linear
activation-function of the hidden units (RLU), and the form of the conditionals
were fixed. Three hyperparameters were crossvalidated using grid-search: the number
of components on each one-dimensional conditional was chosen from the set
$\{2,5,10,20\}$; the weight-decay (used only to regularize the input to hidden weights) from the set $\left\{2.0, 1.0, 0.1, 0.01, 0.001, 0\right\}$;
and the learning rate from the set $\left\{ 0.1, 0.05, 0.025, 0.0125\right\}$.
Learning-rates were decreased linearly to reach 0 after the last epoch.

We also trained fully-visible Bayesian networks (FVBN), an autoregressive model
where each one-dimensional conditional is modelled by a separate mixture
density network using no parameter tying. The same cross-validation procedure
and hyperparameters as for RNADE training were used.
The best validationwise MDN for each one-dimensional conditional was chosen.

The results are shown in Table~\ref{tab:UCI-results}. Autoregressive methods
obtained statistical performances superior to mixture models on all datasets. An
RNADE with mixture of Gaussian conditionals was among the statistically
significant group of best models on all datasets.
Unfortunately we could not reproduce the data-folds used by previous work,
however, our improvements are larger than those demonstrated by a deep mixture
of factor analyzers over standard MFA~\citep{Tang2012}.

\subsection{Natural image patches}
\begin{figure}
\centerline{\includegraphics[width=0.85\textwidth]{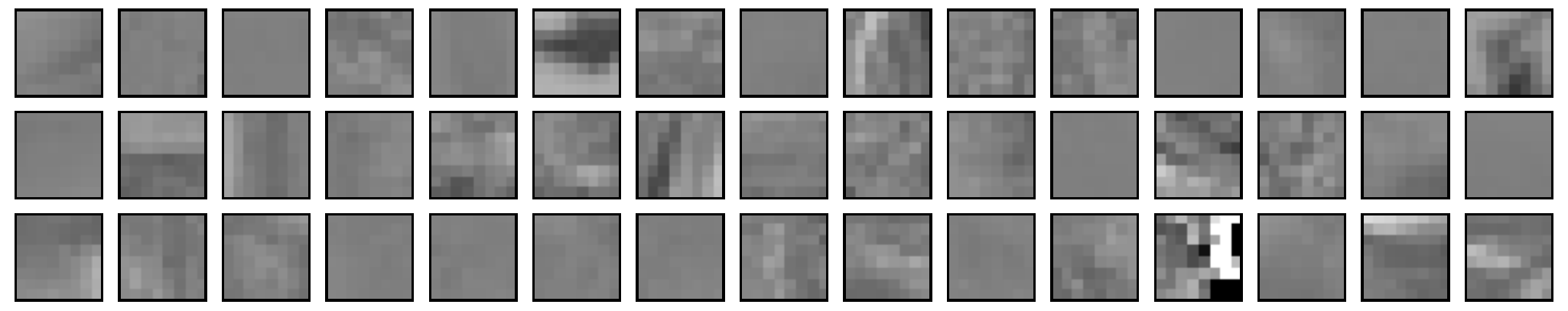}}
\vspace*{-1ex}
\caption{\textbf{Top:}~15 8x8 patches from the BSDS test set. \textbf{Center:}~15 samples
from Zoran and Weiss's MoG model with 200 components. \textbf{Bottom:}~15 samples from
an RNADE with 512 hidden units and 10 output components per dimension. All data
and samples were drawn randomly.}
\label{fig:natimg-samples}
\end{figure}

We also measured the ability of RNADE to model small patches of natural images.
Following the recent work of \citet{Zoran2011},
we use 8-by-8-pixel patches of monochrome natural
images, obtained from the BSDS300 dataset~\citep{Martin2001}
(Figure~\ref{fig:natimg-samples} gives examples).

Pixels in this dataset can take a finite number of brightness values ranging
from $0$ to~$255$. Modeling discretized data using a real-valued distribution
can lead to arbitrarily high density values, by locating narrow high density
spike on each of the possible discrete values. In order to avoid this `cheating'
solution, we added noise uniformly distributed between $0$ and~$1$ to the value
of each pixel. We then divided by~$256$, making each pixel take a value in the
range~$[0,1]$.

In previous experiments, \citet{Zoran2011} subtracted the mean pixel
value from each patch, reducing the dimensionality of
the data by one: the value of any pixel could be perfectly
predicted as minus the sum of all other pixel values. However, the original
study still used a mixture of full-covariance 64-dimensional Gaussians. Such a
model could obtain arbitrarily high model likelihoods, so unfortunately
the likelihoods reported in previous work on this
dataset~\citep{Zoran2011,Tang2012} are difficult to interpret.
In our preliminary experiment using RNADE, we observed that if we model the
64-dimensional data, the 64th pixel is always predicted by a very thin spike
centered at its true value. The ability of RNADE to capture this spurious
dependency is reassuring, but we wouldn't want our results to be dominated by
it.
Recent work by \citet{Zoran2012}, projects the data on the leading 63
eigenvectors of each component, when measuring the model
likelihood~\cite{Zoran2013private}. For comparison amongst a range of methods,
we advocate simply discarding the 64th (bottom-right) pixel.

We trained our model using patches drawn randomly from 180 images in the
training subset of BSDS300. A validation dataset containing 1,000 random patches
from the remaining 20 images in the training subset were used for early-stopping
when training RNADE\@. We measured the performance of each model by measuring
their log-likelihood on one million patches drawn randomly from the test subset,
which is composed of 100 images not present in the training subset.
Given the larger scale of this dataset, hyperparameters of the RNADE and MoG
models were chosen manually using the performance of preliminary runs on the
validation data, rather than by an extensive search.

The RNADE model had 512 rectified-linear hidden units and a mixture of 20
one-dimensional Gaussian components per output. Training was done by
minibatch gradient descent, with 25 datapoints per minibatch, for a total of 200
epochs, each comprising 1,000 minibatches. The learning-rate was scheduled to
start at 0.001 and linearly decreased to reach 0 after the last epoch. Gradient
momentum with momentum factor 0.9 was used, but initiated at the beginning of
the second epoch. A weight decay rate of 0.001 was applied to the
input-to-hidden weight matrix only. Again, we found that multiplying the
gradient of the mean output parameters by the standard deviation improves
results. RNADE training was early stopped but didn't show signs of overfitting.
We produced a further run with 1024 hidden units for 400 epochs, with still no
signs of overfitting; even larger models might perform better.

The MoG model was trained using minibatch EM, for 1,000 iterations. At each
iteration 20,000 randomly sampled datapoints were used in an EM update. A step
was taken from the previous mixture model towards the parameters resulting from
the M-step: $\vec{\theta_{t}} =
(1-\eta)\vec{\theta_{t-1}}+\eta\vec{\theta_{EM}}$, where the step size ($\eta$)
was scheduled to start at 0.1 and linearly decreased to reach~0 after the last
update. The training of the MoG was also early-stopped and also showed no signs
of overfitting.

The results are shown in Table~\ref{tab:BSDS-results}. We compare RNADE with a
mixtures of Gaussians model trained on 63 pixels, and with a MoG trained by
Zoran and Weiss (downloaded from Daniel Zoran's website) from which we removed the
64th row and column of each covariance matrix.
The best RNADE test log-likelihood is, on average, 0.7\;nats per patch lower than
Zoran and Weiss's MoG, which had a different training procedure than our mixture
of Gaussians.

\begin{table}
\vspace*{-0.1in}
\begin{center}
\caption{Average per-example log-likelihood of several mixture of Gaussian and
RNADE models, with mixture of Gaussian (MoG) or mixture of Laplace (MoL) conditionals, on 8-by-8
patches of natural images. These results are measured in nats and were
calculated using one million patches. Standard errors due to the finite test
sample size are lower than~$0.1$ in every case. $K$ gives the number of one-dimensional components for each
conditional in RNADE, and the number of full-covariance components for MoG\@.}
\label{tab:BSDS-results}
\medskip
\begin{tabular}{lSc}
\toprule
{Model} & {Training LogL} & {Test LogL}\\
\midrule
MoG $K\!=\!200$ (Z\&W) & 161.9 & \textbf{152.8}\\
MoG $K\!=\!100$ & 152.8 & 144.7\\
MoG $K\!=\!200$ & 159.3 & 150.4 \\
MoG $K\!=\!300$ & 159.3 & 150.4\\
RNADE-MoG $K\!=\!5$ & 158.0 & 149.1\\
RNADE-MoG $K\!=\!10$& 160.0 & 151.0\\
RNADE-MoG $K\!=\!20$ & 158.6 & 149.7 \\
RNADE-MoL $K\!=\!5$ & 150.2 & 141.5 \\
RNADE-MoL $K\!=\!10$ & 149.7 & 141.1  \\
RNADE-MoL $K\!=\!20$ & 150.1 & 141.5 \\
RNADE-MoG $K\!=\!10$ (sigmoid h.\ units) & 155.1 & 146.4\\
RNADE-MoG $K\!=\!10$ (1024 units, 400 epochs) & 161.1 & 152.1 \\
\bottomrule
\end{tabular}
\end{center}
\vspace*{-0.05in}
\end{table}

Figure~\ref{fig:natimg-samples} shows a few examples from the test set, and
samples from the MoG and RNADE models. Some of the samples from RNADE are
unnaturally noisy, with pixel values outside the legal range (see fourth sample
from the right in Figure~\ref{fig:natimg-samples}). If we constrain the pixels
values to a unit range, by rejection sampling or otherwise, these artifacts go away. Limiting the output range of the model would also improve
test likelihood scores slightly, but not by much: log-likelihood does not
strongly penalize models for putting a small fraction of probability mass on
`junk' images.

All of the results in this section were obtained by fitting the pixels in a
raster-scan order. Perhaps surprisingly, but consistent with previous results
on NADE~\citep{Larochelle2011} and by \citet{frey1998}, randomizing the order of the pixels made little
difference to these results. The difference in performance was comparable to
the differences between multiple runs with the same pixel ordering.

\subsection{Speech acoustics}

We also measured the ability of RNADE to model small patches of speech
spectrograms, extracted from the TIMIT dataset~\cite{Garofolo1993}. The patches
contained 11 frames of 20 filter-banks plus energy; totaling 231 dimensions
per datapoint. These filter-bank encoding is common in
speech-recognition, and better for visualization than the more frequently used MFCC features. A good generative
model of speech
could be used,
for example,
in denoising, or speech detection tasks.

We fitted the models using the standard TIMIT training subset, and
compared RNADE with a MoG by measuring their log-likelihood on the complete TIMIT
core-test dataset.

The RNADE model has 1024 rectified-linear hidden units and a mixture of 20
one-dimensional Gaussian components per output. Given the larger scale of this
dataset hyperparameter choices were again made manually using validation data,
and the same minibatch training procedures for RNADE and MoG were used as for
natural image patches.

\begin{figure}[!t]
\centerline{\includegraphics[width=0.85\textwidth]{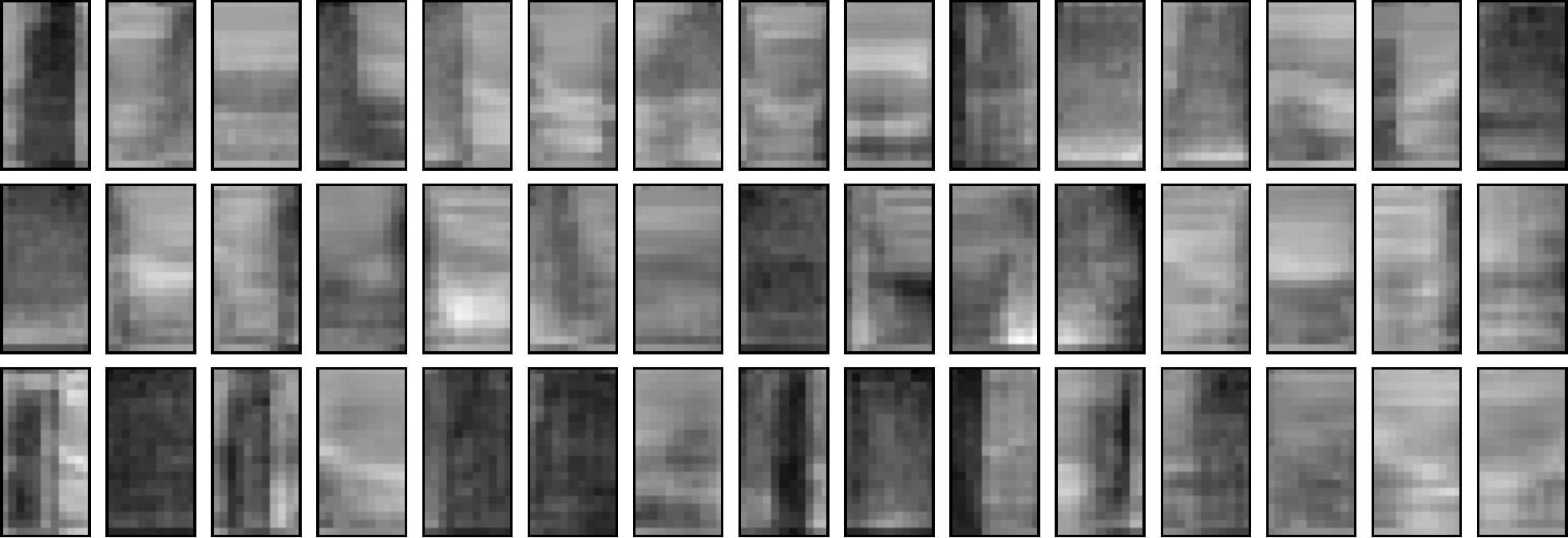}}
\vspace*{-1ex}
\caption{\textbf{Top:} 15 datapoints from the TIMIT core-test set. \textbf{Center:} 15 samples
from a MoG model with 200 components. \textbf{Bottom:}~15 samples from an RNADE
with 1024 hidden units and output components per dimension. On each plot,
time is shown on the horizontal axis, the bottom row displays the energy feature, while the
others display the filter bank features (in ascending frequency order from the
bottom). All data and samples were drawn randomly.}

\label{fig:speech-samples}
\end{figure}

The results are shown in Table \ref{tab:TIMIT-results}. RNADE obtained, on
average, 10 nats more per test example than a mixture of Gaussians. In Figure
\ref{fig:speech-samples} a few examples from the test set, and samples from the
MoG and RNADE models are shown. In contrast with the log-likelihood measure,
there are no marked differences between the samples from each model. Both set of
samples look like blurred spectrograms, but RNADE seems to capture sharper
formant structures (peaks of energy at the lower frequency bands
characteristic of vowel sounds).

\begin{table}[!t]
\begin{center}
\caption{Log-likelihood of several MoG and RNADE models on the core-test set of
TIMIT measured in nats. Standard errors due to the finite test sample size are
lower than $0.3$ nats in every case. RNADE obtained a higher (better)
log-likelihood.}
\label{tab:TIMIT-results}
\medskip
\begin{tabular}{lcc}
\toprule
{Model} & {Training LogL} & {Test LogL} \\
\midrule
MoG $N\!=\!50$  & 111.6 & 110.4\\
MoG $N\!=\!100$ & 113.4 & 112.0\\
MoG $N\!=\!200$ & 113.9 & 112.5\\
MoG $N\!=\!300$ & 114.1 & 112.5\\
RNADE-MoG $K\!=\!10$ & 125.9 & 123.9\\
RNADE-MoG $K\!=\!20$ & 126.7 & \textbf{124.5}\\
RNADE-MoL $K\!=\!10$ & 120.3 & 118.0\\
RNADE-MoL $K\!=\!20$ & 122.2 & 119.8\\
\bottomrule
\end{tabular}
\end{center}
\end{table}

\section{Discussion}
\label{sec:related}

Mixture Density Networks (MDNs)~\citep{Bishop1994} are a flexible conditional
model of probability densities, that can capture skewed, heavy-tailed, and
multi-modal distributions. In principle, MDNs can be applied to
multi-dimensional data. However, the number of parameters that the network has
to output grows quadratically with the number of targets, unless the targets are
assumed independent. RNADE exploits an autoregressive framework to apply
practical, one-dimensional MDNs to unsupervised density estimation.

\begin{figure}
\centerline{\includegraphics[width=\textwidth]{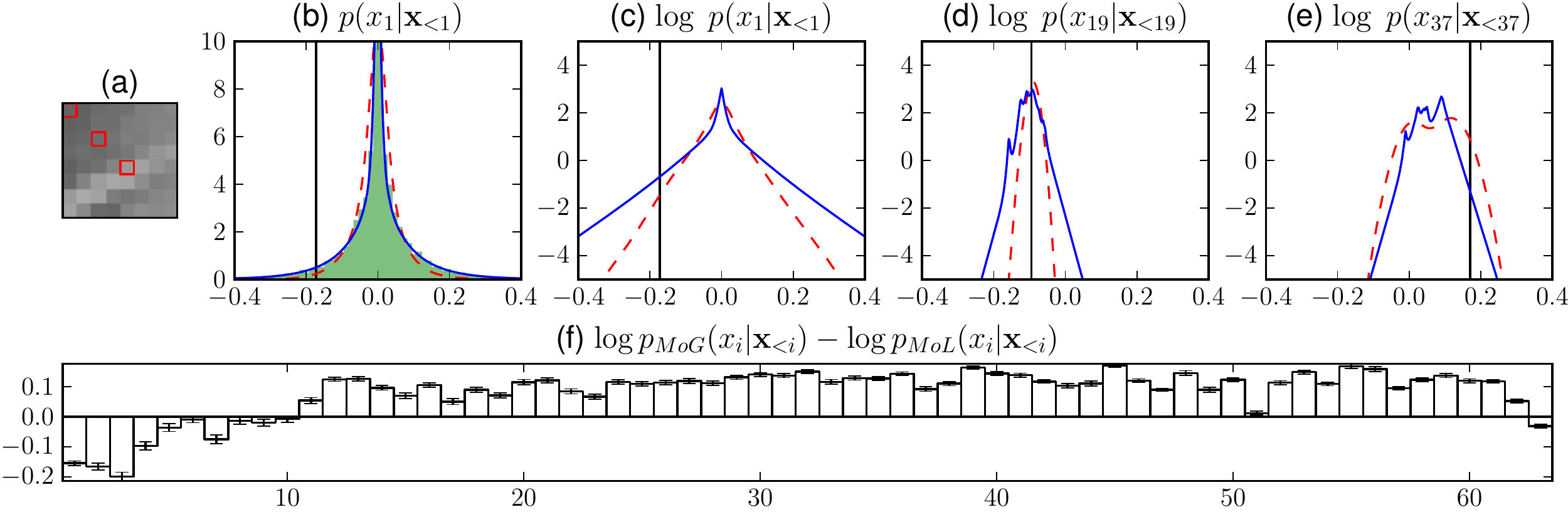}}
\vspace*{-1ex}
\caption{Comparison of Mixture of Gaussian (MoG) and Mixture of Laplace (MoL) conditionals.
\textbf{(a)}~Example test patch.
\textbf{(b)}~Density of $p(x_1)$
under RNADE-MoG (dashed-red) and RNADE-MoL
(solid-blue), both with $K\!=\!10$.
RNADE-MoL closely matches a histogram of brightness values from patches in the test-set (green).
The vertical line indicates the value in~(a).
\textbf{(c)}~Log-density of the distributions in~(b).
\textbf{(d)}~Log-density of MoG and MoL conditionals of pixel 19 in (a).
\textbf{(e)}~Log-density of MoG and MoL conditionals of pixel 37 in (a).
\textbf{(f)}~Difference in predictive log-density between MoG and MoL
conditionals for each pixel, averaged over 10,000 test patches.
}
\label{fig:mol-vs-mog}
\end{figure}

To specify an RNADE we needed to set the parametric form for the output
distribution of each MDN\@. A sufficiently large mixture of Gaussians can closely represent
any density, but it is hard to learn the conditional densities found
in some problems with this representation.
The marginal for the brightness of a pixel in natural image patches is heavy
tailed, closer to a Laplace distribution than Gaussian. Therefore, RNADE-MoG
must fit predictions of the first pixel, $p(x_1)$, with several Gaussians of
different widths, that coincidentally have zero mean. This solution can be
difficult to fit, and RNADE with a mixture of Laplace outputs predicted the
first pixel of image patches better than with a mixture of Gaussians
(Figure~\ref{fig:mol-vs-mog}b and~c).
However, later pixels were predicted better with Gaussian outputs
(Figure~\ref{fig:mol-vs-mog}f); the mixture of Laplace model is not suitable for
predicting with large contexts. For image patches, a scale mixture can work
well~\citep{Theis2012a}, and could be explored within our framework. However for
general applications, scale mixtures within RNADE would be too restrictive (e.g.,
$p(x_1)$ would be zero-mean and unimodal). More flexible one-dimensional forms may aid
RNADE to generalize better for different context sizes and across a range of
applications.

One of the main drawbacks of RNADE, and of neural networks in general, is the
need to decide the value of several training hyperparameters.
The gradient
descent learning rate can be adjusted automatically using, for example, the
techniques developed by \citet{Schaul2013}.
Also, methods for choosing hyperparameters more efficiently than grid search
have been recently developed~\citep{Bergstra2012,Snoek2012}.
These, and several other recent improvements in the neural network field, like
dropouts~\cite{Hinton2012}, should be directly applicable to RNADE, and possibly
obtain even better performance than shown in this work. RNADE makes it
relatively straight-forward to translate advances in the neural-network field
into better density estimators, or at least into new estimators with different
inductive biases.

In summary, we have presented RNADE, a novel `black-box' density estimator. Both
likelihood computation time and the number of parameters scale linearly with the
dataset dimensionality. Generalization across a range of tasks, representing
arbitrary feature vectors, image patches, and auditory spectrograms is excellent.
Performance on image patches was close to a recently reported state-of-the-art
mixture model~\citep{Zoran2011}, and RNADE outperformed mixture models on all other
datasets considered.

\subsubsection*{Acknowledgments}
\ifnipsfinal
We thank John Bridle, Steve Renals, Amos Storkey, and Daniel Zoran for useful interactions.
\else
We thank XXX, YYY and ZZZ for useful discussions.
\fi

\appendix\section{Implementation details}

In this appendix we provide pseudo-code for the calculation of densities and
learning gradients. No new material is presented. A Python implementation of
RNADE is available from
\href{http://www.benignouria.com/en/research/RNADE}{http://www.benignouria.com/en/research/RNADE}.

\begin{algorithm}
\begin{algorithmic}
\caption{Computation of $p(\vec{x})$}
\label{alg:density}
    \State $\vec{a} \gets \vec{c}$
    \State $p(\vec{x}) \gets 1$
    \For{$d$ from 1 to $D$}
        \State $\vec{\psi}_d \gets \rho_d \vec{a}$
        \Comment Rescaling factors
        \State $\hd \gets \vec{\psi}_d\;\mathbf{1}_{\vec{\psi}_d > 0}$
        \Comment Rectified linear units
        \State $\vec{z_d^{\alpha}} \gets {\mat{V}_d^\alpha}\transp\hd +
        \vec{b}_d^\alpha$ \State $\vec{z_d^{\mu}} \gets
        {\mat{V}_d^\mu}\transp\hd + \vec{b}_d^\mu$
        \State $\vec{z_d^{\sigma}} \gets {\mat{V}_d^\sigma}\transp\hd +
        \vec{b}_d^\sigma$
        \State $\vec{\alpha}_d \gets \softmax(\vec{z}_d^{\alpha})$
        \Comment Enforce constraints
        \State $\vec{\mu}_d  \gets \vec{z}_d^{\mu}$
        \State $\vec{\sigma}_d  \gets \exp(\vec{z}_d^{\sigma})$
        \State $p(\vec{x}) \gets p(\vec{x})p_{MoG}(x_d ; \vec{\alpha}_d,
        \vec{\mu}_d, \vec{\sigma}_d)$
        \Comment $p_{MoG}$ is the density of a mixture of Gaussians
        \State $\vec{a} \gets \vec{a} + x_d \mat{W}_{\cdot,d}$
        \Comment Activations are calculated recursively, $x_d$ is a scalar
    \EndFor

    \Return $p(\vec{x})$
\end{algorithmic}
\end{algorithm}

\begin{algorithm}
\caption{Computation of the learning gradients for a datapoint $\vec{x}$}
\label{alg:gradients}
\begin{algorithmic}
    \State $\vec{a} \gets \vec{c}$
    \For{$d$ from $1$ to $D$}
    \Comment Compute the activation of the last dimension
        \State $\vec{a} \gets \vec{a} + x_d \mat{W}_{\cdot,d}$
    \EndFor
    \For{$d$ from $D$ to $1$}
    \Comment Backpropagate errors
        \State $\vec{\psi} \gets \rho_d \vec{a}$
        \Comment Rescaling factors
        \State $\vec{h} \gets \vec{\psi}\;\mathbf{1}_{\vec{\psi} > 0}$
        \Comment Rectified linear units
        \State $\vec{z^{\alpha}} \gets {\mat{V}_d^\alpha}\transp\vec{h} +
        \vec{b}_d^\alpha$
        \State $\vec{z^{\mu}} \gets
        {\mat{V}_d^\mu}\transp\vec{h} + \vec{b}_d^\mu$
        \State $\vec{z^{\sigma}} \gets {\mat{V}_d^\sigma}\transp\hd +
        \vec{b}_d^\sigma$
        \State $\vec{\alpha} \gets \softmax(\vec{z}^{\alpha})$
        \Comment Enforce constraints
        \State $\vec{\mu} \gets \vec{z}^{\mu}$
        \State $\vec{\sigma} \gets \exp(\vec{z}^{\sigma})$

        \State $\vec{\phi} \gets \frac{1}{2}
        \frac{(\vec{\mu}-\vec{x}_d)^2}{\vec{\sigma}^2} - \log \vec{\sigma}
        -\frac{1}{2}\log(2\pi) $
        \Comment Calculate gradients

        \State $\vec{\pi} \gets \frac{\vec{\alpha}\phi}{\sum_{j=1}^K
        \vec{\alpha}_{j}\phi_j}$

        \State $\partial z^{\alpha} \gets \vec{\pi} - \vec{\alpha} $
        \State $\partial\mat{V}^{\alpha}_d \gets \partial z^{\alpha} \vec{h}$
        \State $\partial\mat{b}^{\alpha}_d \gets \partial z^{\alpha}$

        \State $\partial z^{\mu} \gets
        \vec{\pi} (x_d - \vec{\mu}) / \vec{\sigma}^2 $
        \State $\partial z^{\mu} \gets \partial z^{\mu} * \sigma $
        \Comment Move tighter components slower, allows higher learning rates
        \State $\partial\mat{V}^{\mu}_d \gets \partial z^{\mu} \vec{h}$
        \State $\partial\mat{b}^{\mu}_d \gets \partial z^{\mu} $

        \State $\partial z^{\sigma} \gets
        \vec{\pi} \{(x_d - \vec{\mu})^2 / \vec{\sigma}^2 -1\}$
        \State $\partial\mat{V}^{\sigma}_d \gets \partial z^{\sigma}\vec{h}$
        \State $\partial\mat{b}^{\sigma}_d \gets \partial z^{\sigma}$

        \State $\partial\vec{h} \gets \partial\vec{z}^{\alpha} \mat{V}_d^\alpha +
        \partial\vec{z}^{\mu} \mat{V}_d^\mu +
        \partial\vec{z}^{\sigma} \mat{V}_d^\sigma $

        \State $\partial\vec{\psi} \gets \partial\vec{h}\mathbf{1}_{\vec{\psi} > 0}$
        \Comment Second factor: indicator function with condition $\vec{\psi} > 0$
        \State $\partial\rho_d \gets \sum_j \partial\vec{\psi}_j a_j $
        \State $\partial\vec{a} \gets \partial\vec{a} + \partial\vec{\psi} \rho $
        \State $\partial\mat{W}_{\cdot,d} \gets \partial\vec{a} x_d$
        \If{$d=1$}
            \State $\partial\mat{c} \gets \partial\vec{a}$
        \Else
            \State $\vec{a} \gets \vec{a} - x_d \mat{W}_{\cdot,d}$
        \EndIf
    \EndFor

\Return $\partial\vec{\rho}$,
            $\partial\mat{W}$,
            $\partial\vec{c}$,
            $\partial\mat{b}^{\alpha}$,
            $\partial\mat{V}^{\alpha}$,
            $\partial\mat{b}^{\mu}$,
            $\partial\mat{V}^{\mu}$,
            $\partial\mat{b}^{\sigma}$,
            $\partial\mat{V}^{\sigma}$
\end{algorithmic}
\end{algorithm}

In Algorithm~\ref{alg:density} we detail the pseudocode for
calculating the density of a datapoint under an RNADE with mixture of Gaussian
conditionals. The model has parameters:
$\vec{\rho} \in \realdomain^{D}$,
$\mat{W} \in \realdomain^{H \times D-1}$,
$\vec{c} \in \realdomain^{H}$,
$\mat{b}^{\alpha} \in \realdomain^{D \times K}$,
$\mat{V}^{\alpha} \in \realdomain^{D \times H \times K}$,
$\mat{b}^{\mu} \in \realdomain^{D \times K}$,
$\mat{V}^{\mu} \in \realdomain^{D \times H \times K}$,
$\mat{b}^{\sigma} \in \realdomain^{D \times K}$,
$\mat{V}^{\sigma} \in \realdomain^{D \times H \times K}$

Training of an RNADE model can be done using a gradient ascent algorithm on the
log-likelihood of the model given the training data. Gradients can be calculated
using automatic differentiation libraries (\eg Theano~\cite{bergstra2010}). However we found our manual implementation to work
faster in practice, possibly due to our recomputation of the $\vec{a}$ terms in
the second \emph{for} loop in Algorithm~\ref{alg:gradients}, which is more
cache-friendly than storing them during the first loop.

Here we show the derivation of the gradients of each parameter of a NADE model
with MoG conditionals.
Following \cite{Bishop1994}, we define $\phi_i(x_d \given \xltd)$ as the density
of $x_d$ under the $i$-th component of the conditional:
\begin{equation}
\phi_i(x_d \given \xltd) = \frac{1}{\sqrt{2
\pi}\vec{\sigma}_{d,i}}\exp\left\{-\frac{(x_d-\vec{\mu}_{d,i})^2}{2\vec{\sigma}_{d,i}^2}
\right\}\mathrm{,}
\end{equation}
and $\pi_i(x_d \given \xltd)$ as the ``responsibility'' of the $i$-th component
for $x_d$:
\begin{equation}
\pi_i(x_d \given \xltd) = \dfrac{ \vec{\alpha}_{d,i}\phi_i(x_d \given \xltd)  }{
\sum_{j=1}^K \vec{\alpha}_{d,j}\phi_j(x_d \given \xltd)}\mathrm{.}
\end{equation}
It is easy to find just by taking their derivatives that:
\begin{align}
\dfrac{\partial p(\vec{x})}{\partial \vec{z}_{d,i}^{\alpha}} & =
\pi_i(x_d \given \xltd) - \vec{\alpha}_{d,i} \\
\dfrac{\partial p(\vec{x})}{\partial \vec{z}_{d,i}^{\mu}} & =
\pi_i(x_d \given \xltd) \dfrac{x_d-\vec{\mu}_{d,i}}{\vec{\sigma}_{d,i}^2} \\
\dfrac{\partial p(\vec{x})}{\partial \vec{z}_{d,i}^{\sigma}} & =
\pi_i(x_d \given \xltd) \left\{
\dfrac{(x_d-\vec{\mu}_{d,i})^2}{\vec{\sigma}_{d,i}^2} -1\right\}
\end{align}

Using the chain rule we can calculate the derivative of the
parameters of the output layer parameters:
\begin{align}
\dfrac{\partial p(\vec{x})}{\partial \mat{V}_d^\alpha} & =
\dfrac{\partial p(\vec{x})}{\partial \vec{z}_{d,i}^{\alpha}}
\dfrac{\partial \vec{z}_{d,i}^{\alpha}}{\mat{V}_d^\alpha} =
\dfrac{\partial p(\vec{x})}{\partial \vec{z}_{d,i}^{\alpha}} \vec{h} \\
\dfrac{\partial p(\vec{x})}{\partial \mat{b}_d^\alpha} & =
\dfrac{\partial p(\vec{x})}{\partial \vec{z}_{d,i}^{\alpha}}
\dfrac{\partial \vec{z}_{d,i}^{\alpha}}{\mat{b}_d^\alpha} =
\dfrac{\partial p(\vec{x})}{\partial \vec{z}_{d,i}^{\alpha}} \\
\dfrac{\partial p(\vec{x})}{\partial \mat{V}_d^\mu} & =
\dfrac{\partial p(\vec{x})}{\partial \vec{z}_{d,i}^{\mu}}
\dfrac{\partial \vec{z}_{d,i}^{\alpha}}{\mat{V}_d^\mu} =
\dfrac{\partial p(\vec{x})}{\partial \vec{z}_{d,i}^{\mu}} \vec{h} \\
\dfrac{\partial p(\vec{x})}{\partial \mat{b}_d^\mu} & =
\dfrac{\partial p(\vec{x})}{\partial \vec{z}_{d,i}^{\mu}}
\dfrac{\partial \vec{z}_{d,i}^{\alpha}}{\mat{b}_d^\mu} =
\dfrac{\partial p(\vec{x})}{\partial \vec{z}_{d,i}^{\mu}} \\
\dfrac{\partial p(\vec{x})}{\partial \mat{V}_d^\sigma} & =
\dfrac{\partial p(\vec{x})}{\partial \vec{z}_{d,i}^{\sigma}}
\dfrac{\partial \vec{z}_{d,i}^{\alpha}}{\mat{V}_d^\sigma} =
\dfrac{\partial p(\vec{x})}{\partial \vec{z}_{d,i}^{\sigma}} \vec{h} \\
\dfrac{\partial p(\vec{x})}{\partial \mat{b}_d^\sigma} & =
\dfrac{\partial p(\vec{x})}{\partial \vec{z}_{d,i}^{\sigma}}
\dfrac{\partial \vec{z}_{d,i}^{\alpha}}{\mat{b}_d^\sigma} =
\dfrac{\partial p(\vec{x})}{\partial \vec{z}_{d,i}^{\sigma}}
\end{align}

By ``backpropagating'' the we can calculate the partial derivatives with
respect to the output of the hidden units:

\begin{align}
\dfrac{\partial p(\vec{x})}{\partial \vec{h}_d} & =
\dfrac{\partial p(\vec{x})}{\partial \vec{z}_{d,i}^{\alpha}}
\dfrac{\partial \vec{z}_{d,i}^{\alpha}}{\partial \vec{h}_d} +
\dfrac{\partial p(\vec{x})}{\partial \vec{z}_{d,i}^{\mu}}
\dfrac{\partial \vec{z}_{d,i}^{\mu}}{\partial \vec{h}_d} +
\dfrac{\partial p(\vec{x})}{\partial \vec{z}_{d,i}^{\sigma}}
\dfrac{\partial \vec{z}_{d,i}^{\sigma}}{\partial \vec{h}_d} \\
 & =
\dfrac{\partial p(\vec{x})}{\partial \vec{z}_{d,i}^{\alpha}}
 \mat{V}_d^\alpha +
\dfrac{\partial p(\vec{x})}{\partial \vec{z}_{d,i}^{\mu}}
\mat{V}_d^\mu +
\dfrac{\partial p(\vec{x})}{\partial \vec{z}_{d,i}^{\sigma}}
\mat{V}_d^\sigma
\end{align}

and calculate the partial derivatives with respect to all other parameters
in RNADE:
\begin{align}
\dfrac{\partial p(\vec{x})}{\partial \vec{\psi}_d}  & =
\dfrac{\partial p(\vec{x})}{\partial \vec{h}_d} \mathbf{1}_{\vec{\psi}_d > 0} \\
\dfrac{\partial p(\vec{x})}{\partial \rho_d}  & =
\sum_j \dfrac{\partial p(\vec{x})}{\partial \vec{\psi}_{d,j}}  \vec{a}_{d,j} \\
\dfrac{\partial p(\vec{x})}{\partial \vec{a}_d}  & =
\dfrac{\partial p(\vec{x})}{\partial \vec{a}_{d+1}} +
\dfrac{\partial p(\vec{x})}{\partial \vec{h}_d} \rho_d \mathbf{1}_{\vec{\psi}_d
> 0} \label{eq:gradient-act}\\
\dfrac{\partial p(\vec{x})}{\partial \mat{W}_{\cdot,d}} & =
\dfrac{\partial p(\vec{x})}{\partial \vec{a}_d} x_d \\
\dfrac{\partial p(\vec{x})}{\partial \vec{c}} & =
\dfrac{\partial p(\vec{x})}{\partial \vec{a}_1}
\end{align}
Note that gradients are calculated recursively, due to
\eqref{eq:gradient-act}, starting at $d=D$ and progressing down to $d=1$.

\bibliographystyle{unsrtnat}
\bibliography{references}

\begin{thebibliography}{34}
\providecommand{\natexlab}[1]{#1}
\providecommand{\url}[1]{\texttt{#1}}
\expandafter\ifx\csname urlstyle\endcsname\relax
  \providecommand{\doi}[1]{doi: #1}\else
  \providecommand{\doi}{doi: \begingroup \urlstyle{rm}\Url}\fi

\bibitem[Koller and Friedman(2009)]{koller2009}
D.~Koller and N.~Friedman.
\newblock \emph{Probabilistic graphical models: principles and techniques}.
\newblock MIT Press, 2009.

\bibitem[Cacoullos(1966)]{cacoullos1966}
T.~Cacoullos.
\newblock Estimation of a multivariate density.
\newblock \emph{Annals of the Institute of Statistical Mathematics},
  18\penalty0 (1):\penalty0 179--189, 1966.

\bibitem[Zoran and Weiss(2011)]{Zoran2011}
D.~Zoran and Y.~Weiss.
\newblock From learning models of natural image patches to whole image
  restoration.
\newblock In \emph{International Conference on Computer Vision}, pages
  479--486. IEEE, 2011.

\bibitem[Salakhutdinov and Murray(2008)]{Salakhutdinov2008a}
R.~Salakhutdinov and I.~Murray.
\newblock On the quantitative analysis of deep belief networks.
\newblock In \emph{Proceedings of the 25th International Conference on Machine
  learning}, pages 872--879. Omnipress, 2008.

\bibitem[Larochelle and Murray(2011)]{Larochelle2011}
H.~Larochelle and I.~Murray.
\newblock The neural autoregressive distribution estimator.
\newblock \emph{Journal of Machine Learning Research W\&CP}, 15:\penalty0
  29--37, 2011.

\bibitem[Bishop(1994)]{Bishop1994}
C.~M. Bishop.
\newblock Mixture density networks.
\newblock Technical Report NCRG 4288, Neural Computing Research Group, Aston
  University, Birmingham, 1994.

\bibitem[Frey et~al.(1996)Frey, Hinton, and Dayan]{Frey1996}
B.~J. Frey, G.~E. Hinton, and P.~Dayan.
\newblock Does the wake-sleep algorithm produce good density estimators?
\newblock In \emph{Advances in Neural Information Processing Systems 8}, pages
  661--670. MIT Press, 1996.

\bibitem[Bengio and Bengio(2000)]{Bengio2000}
Y.~Bengio and S.~Bengio.
\newblock Modeling high-dimensional discrete data with multi-layer neural
  networks.
\newblock \emph{Advances in Neural Information Processing Systems},
  12:\penalty0 400--406, 2000.

\bibitem[Larochelle and Lauly(2012)]{larochelleneural}
H.~Larochelle and S.~Lauly.
\newblock A neural autoregressive topic model.
\newblock In \emph{Advances in Neural Information Processing Systems 25}, 2012.

\bibitem[Bengio(2011)]{bengiodiscussion}
Y.~Bengio.
\newblock Discussion of “the neural autoregressive distribution estimator”.
\newblock \emph{Journal of Machine Learning Research W\&CP}, 15:\penalty0
  38--39, 2011.

\bibitem[Theis et~al.(2012)Theis, Hosseini, and Bethge]{Theis2012a}
L.~Theis, R.~Hosseini, and M.~Bethge.
\newblock Mixtures of conditional {G}aussian scale mixtures applied to
  multiscale image representations.
\newblock \emph{PLoS ONE}, 7\penalty0 (7), 2012.
\newblock \doi{10.1371/journal.pone.0039857}.

\bibitem[Friedman and Nachman(2000)]{Friedman2000}
N.~Friedman and I.~Nachman.
\newblock Gaussian process networks.
\newblock In \emph{Proceedings of the Sixteenth Conference on Uncertainty in
  Artificial Intelligence}, pages 211--219. Morgan Kaufmann Publishers Inc.,
  2000.

\bibitem[Murray and Salakhutdinov(2009)]{murray2009}
I.~Murray and R.~Salakhutdinov.
\newblock Evaluating probabilities under high-dimensional latent variable
  models.
\newblock In \emph{Advances in Neural Information Processing Systems 21}, pages
  1137--1144, 2009.

\bibitem[Theis et~al.(2011)Theis, Gerwinn, Sinz, and Bethge]{Theis2011}
L.~Theis, S.~Gerwinn, F.~Sinz, and M.~Bethge.
\newblock In all likelihood, deep belief is not enough.
\newblock \emph{Journal of Machine Learning Research}, 12:\penalty0 3071--3096,
  2011.

\bibitem[Ranzato and Hinton(2010)]{Ranzato2010}
M.~A. Ranzato and G.~E. Hinton.
\newblock Modeling pixel means and covariances using factorized third-order
  {B}oltzmann machines.
\newblock In \emph{Computer Vision and Pattern Recognition}, pages 2551--2558.
  IEEE, 2010.

\bibitem[Courville et~al.(2011)Courville, Bergstra, and Bengio]{Courville2011}
A.~Courville, J.~Bergstra, and Y.~Bengio.
\newblock A spike and slab restricted {B}oltzmann machine.
\newblock \emph{Journal of Machine Learning Research, W\&CP}, 15, 2011.

\bibitem[Nair and Hinton(2010)]{nair2010}
V.~Nair and G.~E. Hinton.
\newblock Rectified linear units improve restricted {B}oltzmann machines.
\newblock In \emph{Proceedings of the 27th International Conference on Machine
  Learning}, pages 807--814. Omnipress, 2010.

\bibitem[Bridle(1989)]{Bridle1989}
J.~S. Bridle.
\newblock Probabilistic interpretation of feedforward classification network
  outputs, with relationships to statistical pattern recognition.
\newblock In \emph{Neuro-computing: algorithms, architectures and
  applications}, pages 227--236. Springer-Verlag, 1989.

\bibitem[Robinson(1994)]{robinson1994}
T.~Robinson.
\newblock {SHORTEN}: simple lossless and near-lossless waveform compression.
\newblock Technical Report CUED/F-INFENG/TR.156, Engineering Department,
  Cambridge University, 1994.

\bibitem[Tang et~al.(2012)Tang, Salakhutdinov, and Hinton]{Tang2012}
Y.~Tang, R.~Salakhutdinov, and G.~Hinton.
\newblock Deep mixtures of factor analysers.
\newblock In \emph{Proceedings of the 29th International Conference on Machine
  Learning}, pages 505--512. Omnipress, 2012.

\bibitem[Zoran and Weiss(2012)]{Zoran2012}
D.~Zoran and Y.~Weiss.
\newblock Natural images, {G}aussian mixtures and dead leaves.
\newblock \emph{Advances in Neural Information Processing Systems},
  25:\penalty0 1745--1753, 2012.

\bibitem[Bache and Lichman(2013)]{Bache+Lichman:2013}
K.~Bache and M.~Lichman.
\newblock {UCI} machine learning repository, 2013.
\newblock http://archive.ics.uci.edu/ml.

\bibitem[Silva et~al.(2011)Silva, Blundell, and Teh]{Silva2011}
R.~Silva, C.~Blundell, and Y.~W. Teh.
\newblock Mixed cumulative distribution networks.
\newblock \emph{Journal of Machine Learning Research W\&CP}, 15:\penalty0
  670--678, 2011.

\bibitem[Ghahramani and Hinton(1996)]{Ghahramani1996}
Z.~Ghahramani and G.~E. Hinton.
\newblock The {EM} algorithm for mixtures of factor analyzers.
\newblock Technical Report CRG-TR-96-1, University of Toronto, 1996.

\bibitem[Verbeek(2005)]{verbeek2005}
J.~Verbeek.
\newblock Mixture of factor analyzers {M}atlab implementation, 2005.
\newblock http://lear.inrialpes.fr/~verbeek/code/.

\bibitem[Martin et~al.(2001)Martin, Fowlkes, Tal, and Malik]{Martin2001}
D.~Martin, C.~Fowlkes, D.~Tal, and J.~Malik.
\newblock A database of human segmented natural images and its application to
  evaluating segmentation algorithms and measuring ecological statistics.
\newblock In \emph{International Conference on Computer Vision}, volume~2,
  pages 416--423. IEEE, July 2001.

\bibitem[Zoran(2013)]{Zoran2013private}
D.~Zoran.
\newblock Personal communication, 2013.

\bibitem[Frey(1998)]{frey1998}
B.~Frey.
\newblock \emph{Graphical models for machine learning and digital
  communication}.
\newblock MIT Press, 1998.

\bibitem[Garofolo et~al.(1993)Garofolo, Lamel, Fisher, Fiscus, Pallett,
  Dahlgren, and Zue]{Garofolo1993}
J.~S. Garofolo, L.~F. Lamel, W.~M. Fisher, J.~G. Fiscus, D.~S. Pallett, N.~L.
  Dahlgren, and V.~Zue.
\newblock Timit acoustic-phonetic continuous speech corpus.
\newblock \emph{Linguistic Data Consortium}, 10\penalty0 (5):\penalty0 0, 1993.

\bibitem[Schaul et~al.(2013)Schaul, Zhang, and LeCun]{Schaul2013}
T.~Schaul, S.~Zhang, and Y.~LeCun.
\newblock {No More Pesky Learning Rates}.
\newblock In \emph{Proceedings of the 30th international conference on Machine
  learning}, 2013.

\bibitem[Bergstra and Bengio(2012)]{Bergstra2012}
J.~Bergstra and Y.~Bengio.
\newblock Random search for hyper-parameter optimization.
\newblock \emph{The Journal of Machine Learning Research}, 13:\penalty0
  281--305, 2012.

\bibitem[Snoek et~al.(2012)Snoek, Larochelle, and Adams]{Snoek2012}
J.~Snoek, H.~Larochelle, and R.~Adams.
\newblock Practical {B}ayesian optimization of machine learning algorithms.
\newblock In \emph{Advances in Neural Information Processing Systems 25}, pages
  2960--2968, 2012.

\bibitem[Hinton et~al.(2012)Hinton, Srivastava, Krizhevsky, Sutskever, and
  Salakhutdinov]{Hinton2012}
G.~E. Hinton, N.~Srivastava, A.~Krizhevsky, I.~Sutskever, and R.~R.
  Salakhutdinov.
\newblock Improving neural networks by preventing co-adaptation of feature
  detectors.
\newblock \emph{Arxiv preprint arXiv:1207.0580}, 2012.

\bibitem[Bergstra et~al.(2010)Bergstra, Breuleux, Bastien, Lamblin, Pascanu,
  Desjardins, Turian, Warde-Farley, and Bengio]{bergstra2010}
James Bergstra, Olivier Breuleux, Fr{\'{e}}d{\'{e}}ric Bastien, Pascal Lamblin,
  Razvan Pascanu, Guillaume Desjardins, Joseph Turian, David Warde-Farley, and
  Yoshua Bengio.
\newblock Theano: a {CPU} and {GPU} math expression compiler.
\newblock In \emph{Proceedings of the Python for Scientific Computing
  Conference ({SciPy})}, June 2010.
\newblock Oral Presentation.

\end{thebibliography}

\end{document}